\PassOptionsToPackage{table}{xcolor}
\documentclass[sigconf,screen]{acmart}

\AtBeginDocument{%
  }

\usepackage{xspace}
\usepackage{array}
\usepackage{xcolor}

\def\abstractname{ABSTRACT}

\def\keywordsname{KEYWORDS}
\def\refname{REFERENCES}

\newcommand{\onedot}{.\xspace}

\def\etal{et al\onedot}
\makeatother

\setcopyright{none}
\renewcommand\footnotetextcopyrightpermission[1]{}
\acmYear{2025}
\acmConference[Conference'25]{}{August, 2025}{}
\begin{document}

\title{GeoTexBuild: 3D Building Model Generation from Map Footprints}

\author{Ruizhe Wang}  
\email{rz\_wang@seu.edu.cn}
\affiliation{%
  \institution{Southeast University}
  \country{}
}

\author{Junyan Yang}
\email{yangjy\_seu@163.com} 
\affiliation{%
  \institution{Southeast University}
  \country{}
}

\author{Qiao Wang}
\email{qiaowang@seu.edu.cn}
\orcid{0000-0002-5271-0472}
\affiliation{%
  \institution{Southeast University}
  \country{}
  % \city{Nanjing}
  % \state{Jiangsu}
  % \country{China}
}
\renewcommand{\shortauthors}{Wang et al.}

\begin{abstract}
  We introduce GeoTexBuild, a modular generative framework for creating 3D building models from footprints derived from site planning or map designs. The system is designed for architects and city planners, offering a seamless solution that directly converts map features into 3D buildings. The proposed framework employs a three-stage process comprising height map generation, geometry reconstruction, and appearance stylization, culminating in building models with detailed geometry and appearance attributes. By integrating customized ControlNet, Neural style field (NSF), and Multi-view diffusion model, we explore effective methods for controlling both geometric and visual attributes during the generation process. Our approach eliminates the problem of structural variations in a single facade image in existing 3D generation techniques for buildings. Experimental results at each stage validate the capability of GeoTexBuild to generate detailed and accurate building models from footprints.
\end{abstract}




\begin{teaserfigure}
  \includegraphics[width=\textwidth]{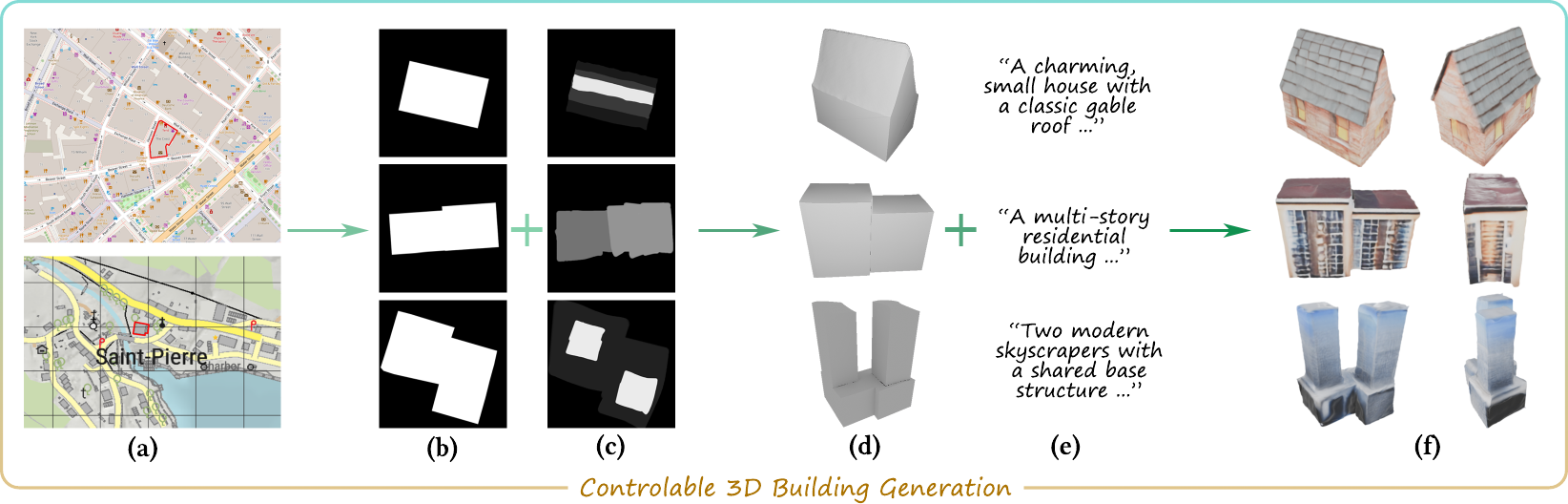}
  \caption{Building3D allows users to generate textured 3D building models (f) from a footprint on maps (b) extracted from (a), a free-hand height sketch (c), and a style prompt (e), enabling precise control on both geometry and visual effects. 
  % Column (a) screenshots from Map data © OpenStreetMap contributors, licensed under ODbL, Arma Reforger © Bohemia Interactive.
  }
  \Description{}
  \label{fig:fpo}
\end{teaserfigure}

\maketitle

% \footnotetext[1]{\{rz\_wang, qiaowang\}@seu.edu.cn, yangjy\_seu@163.com}

\section{INTRODUCTION}
Building models are integral to architectural design, urban planning, tourism, and virtual reality applications \cite{3durban, 3dvirtual}. However, the complexity of architectural structures poses challenges for automated and controllable modeling tailored to specific design needs, making it an area ripe for further exploration \cite{Wei_2023_ICCV}. While manual model creation using CAD software is time-consuming and labor-intensive, modern 3D reconstruction technologies \cite{lin2024vastgaussian, blocknerf, li2023neuralangelo, schoenberger2016sfm}—despite their significant advancements—face limitations: they cannot generate novel buildings beyond existing real-world structures, often require extensive and costly data collection \cite{tao2024oxfordspiresdatasetbenchmarking}, thus are poorly suited for early-stage planning and designing.

Accurate 3D building modeling demands more than a single facade design or street-view imagery due to the complexity of architectural geometric structures \cite{PANG2022102859}. This intricacy extends beyond the facade to include structural design, height, and proportionality. However, existing 3D generation techniques \cite{poole2022dreamfusion, bensadoun2024meta3dgen, chen2023fantasia3d, shi2023MVDream, zhao2025hunyuan3d20scalingdiffusion, tochilkin2024triposrfast3dobject}—which rely on single images or textual prompts—often lack the geometric control required for intricate structures. Consequently, more effective feature representations are needed to encapsulate and regulate a building’s overall geometric characteristics. 

In architectural design, \textbf{(A) Building footprints, identified as natural features on a map (Fig.\ref{fig:fpo} (a)), can serve as effective geometric control parameters.} They are usually established in site planning \cite{lynch1984site} or map designs \cite{moore2016basics} by professional systems and personnel for urban planning and map design. 
\textbf{(B) Once the footprint determines the facade positions, the roof's design (Fig.\ref{fig:fpo} (c)), shaped by variations in its segments and height, reflects the overall complexity of the building's superstructure.}
Thus, we can control its shape by conditioning on the roof structure.

Given these observations, we propose GeoTexBuild (Fig.\ref{fig:pipeline}), a three-stage generative framework comprising height map generation, geometry reconstruction, and appearance stylization. Designed to comply with designer-specified control conditions, GeoTexBuild seamlessly integrates predefined attributes with freehand sketches and textual prompts, enabling the creation of textured architectural models.
First, we integrate ControlNet \cite{zhang2023adding} to fuse building footprint data with hand-drawn height sketches, enabling controlled roof structure generation. Next, the overall building geometry is reconstructed from the generated roof structure and input parameters, yielding an untextured model. Finally, we propose a multi-view mesh stylization approach, leveraging neural style fields \cite{t2m} and multi-view diffusion models \cite{dong2024coin3d} to enhance the model with detailed geometry and color.

In our experiment, we customized multiple ControlNets to evaluate the impact of various image controls on roof structure generation, while constructing a multi-view diffusion-integrated neural style field for building mesh stylization. 
We compare our stylization module with other geometry-controlled stylization methods \cite{t2m, dong2024coin3d} and the framework with text/image to 3D methods \cite{ukarapol2024gradeadreamer, zhao2025hunyuan3d20scalingdiffusion, tochilkin2024triposrfast3dobject}. 
Experiments from various perspectives substantiate the effectiveness of our framework in generating architectural models.
In addition, modules in our framework can be replaced with newer, pre-trained models, allowing us to leverage the latest and future advancements in the rapidly growing fields of vision and language research. 

For applications, we aim to assist architects in quickly progressing from basic urban planning data to more advanced downstream tasks and in transitioning conceptual design to solid models, which support creating static renderings, calculating lighting and shading, or analyzing aesthetic effects within a block. These tasks require a rapid generation of models with certain details, but do not require extreme precision. Our framework represents a step from low-polygon to detailed representation (densification), while previous work either did not start from the perspective of urban planners or was in the opposite direction of ours (simplification).
% GeoTexBuild bridges creative vision and technical execution, streamlining custom architectural design, and can be extended to applications in design for exteriors, interiors, and urban aesthetics.

In summary, our contributions are as follows.
\textbf{(1)}
We present a three-stage modular framework, GeoTexBuild, for generating building models from footprints with detailed controllable geometry and appearance attributes.
\textbf{(2)}
We investigate the control of geometry attributes of building generation by training a customized ControlNet with a variant of the Building3D dataset \cite{Wang2023Building3DAU}.
\textbf{(3)}
We propose a multi-view mesh stylization method based on NSF and MV diffusion models to stylize building meshes.
\textbf{(4)}
Experiments and comparisons with geometry-controlled stylization and text/image conditioned 3D generation methods prove the effectiveness of our framework and each module.

\section{RELATED WORK}
\subsection{Modeling Buildings from Reconstruction} 
In addition to manual modeling, the earliest methods for the automatic modeling of buildings include various approaches to reconstructing structures based on data obtained from reality or the expertise of professionals. 
These approaches typically involve reconstructing buildings from satellite imagery \cite{reconsate, rs11141660}, aerial photography \cite{reconaerial}, multiple photographs \cite{modelingBerkeley, MIASI}, point cloud \cite{PolyFit, chen2022points2poly}, or lidar data \cite{reconlidar1}. 

Semi-automatic modeling based on professional experience often employs procedural modeling techniques \cite{10.1145/383259.383292, 10.1145/1141911.1141931} and methods for generating building grammars from collected data \cite{10.1145/3641519.3657400, recongrammar}. Although these methods are capable of modeling accurate building models, using these grammar-based languages needs extra effort in learning and manual programming; moreover, their structures and styles are in a finite range, limited by grammars and branching logics.

Researchers have developed learning or optimization techniques for reconstruction \cite{schoenberger2016sfm, PolyFit, chen2022points2poly, ProceduralModeling}, updating the computational logic and improving the data volume requirements. 
Furthermore, novel methods have emerged for general 3D reconstruction, extending beyond just buildings \cite{chen2025survey3dgaussiansplatting, 10521791, gao2023nerfneuralradiancefield}. 
Neural reversed rendering and Gaussian splatting-based methods \cite{Huang2DGS2024, li2023neuralangelo, kerbl3Dgaussians, mueller2022instant, barron2022mip, mildenhall2020nerf} redefine 3D data representation and introduce new principles that enhance both the efficiency and fidelity, making them capable of concurrently restoring geometry and color to a photorealistic level.

Although most of these methods do not aim to generate novel designs, they have advanced the comprehension of modeling and generation processes, and have established both methodological frameworks and data foundations for generation.

\begin{figure*}
    \centering
    \includegraphics[width=1.0\linewidth, trim={0 0 0 0}, clip]{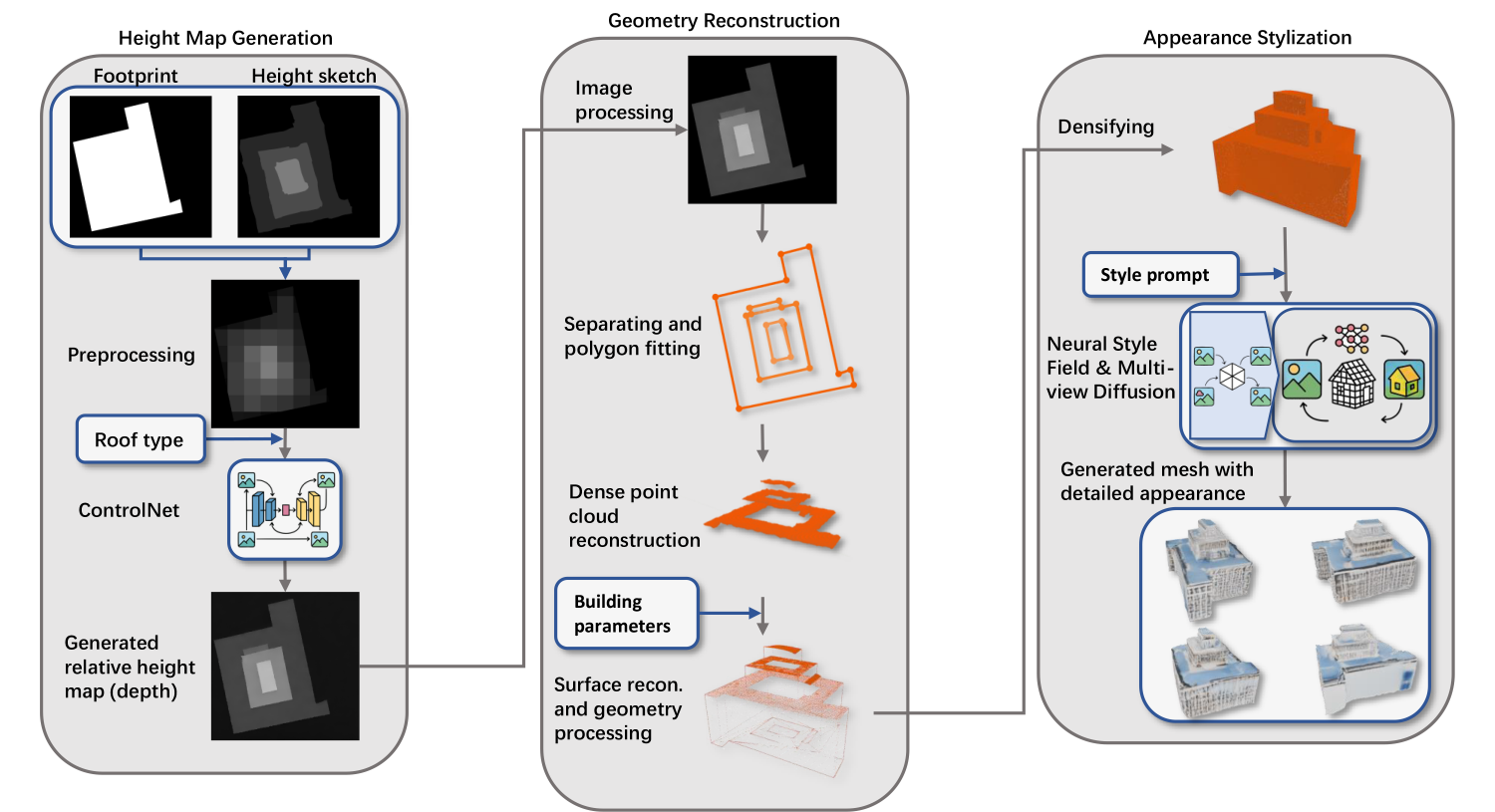}
    \caption{Illustration of our framework. Our framework has three stages: height map generation, geometry reconstruction, and appearance stylization.}
    \Description{}
    \label{fig:pipeline}
\end{figure*}

\subsection{3D Generation}
While general 3D generation technology is not typically focused on architecture, building generation remains a significant application, with building models comprising a substantial portion of various datasets. 
3D generation is primarily categorized into two types: One emphasizing the scene generation \cite{10.1007/978-3-031-72904-1_13, Hwang_2023_CVPR, 10422989, DBLP:journals/corr/abs-2406-09394}, like modeling 3D urban environments \cite{10656934, Xu2024Sketch2SceneAG, shang2024urbanworldurbanworldmodel} without the details of a single building; and the other focusing on generating individual objects.
% Single-object 3D generation can be further divided into two categories: some models are capable of producing only shapes \cite{survey3dgen}, while others can simultaneously generate both shapes and appearances. 

The mainstream of generative methods leverages two kinds of input paradigms: text-to-3D \cite{lee2024textto3d} and image-to-3D \cite{li2024advances3dgenerationsurvey}. Different 3D representations give rise to targeted approaches: methods for explicit representation, such as \cite{nichol2022pointegenerating3dpoint, Wei_2023_ICCV, Zhou_2021_ICCV} for point cloud, \cite{siddiqui2024meta3dassetgentexttomesh, zhao2025hunyuan3d20scalingdiffusion} for mesh, \cite{Ren_2024_CVPR} for voxel, \cite{yi2023gaussiandreamer, ukarapol2024gradeadreamer} for 3DGS; implicit representations, such as \cite{poole2022dreamfusion, raj2023dreambooth3d} for NeRF, \cite{jun2023shapegeneratingconditional3d} for neural implicit surface, \cite{zhao2023michelangelo, Metzer2022LatentNeRFFS} for latent, \cite{hong2024lrm, tochilkin2024triposrfast3dobject} for triplane, \cite{DBLP:journals/corr/abs-2406-09394} for multi-layer representation. 
Regarding generation strategies, models may employ multi-view optimization \cite{tang2023dreamgaussian, shi2023MVDream, ukarapol2024gradeadreamer, yi2023gaussiandreamer, raj2023dreambooth3d, poole2022dreamfusion}, feedforward \cite{Long2023Wonder3DSI, bensadoun2024meta3dgen, shuang2025direct3d, hong2024lrm, tochilkin2024triposrfast3dobject, zhao2025hunyuan3d20scalingdiffusion, yang2025hunyuan3d10unifiedframework}. 
Based on training data, they can be divided into models trained on 3D data \cite{bensadoun2024meta3dgen, shuang2025direct3d, hong2024lrm, tochilkin2024triposrfast3dobject, zhao2025hunyuan3d20scalingdiffusion}, those trained on multi-view 2D images \cite{tang2023dreamgaussian, ukarapol2024gradeadreamer, yi2023gaussiandreamer, raj2023dreambooth3d, poole2022dreamfusion}, and those trained on single-view images \cite{Wei_2023_ICCV}.

Although the SOTA methods can generate plausible models from certain kinds of conditioning, they are trained with tremendous numbers of GPUs and data. Thus, they are not able to adapt to our new control requirements. We outline a suitable technical roadmap for input and output, representations, and generation strategies based on the understanding of these methods, in terms of complexity, training cost, data volume, generation quality, and controllability.
% Classifying 3D generation technologies from multiple perspectives enables us to efficiently outline a suitable technical roadmap for input and output, representations, and generation strategies, as these models exhibit variation in complexity, training cost, required data volume, generation quality, and controllability.

\subsection{3D Model Editing and Stylization}
Incorporating or modifying the appearance of an existing geometric model achieved through editing and appearance stylization techniques is important for 3D creation. 
Certain stylization approaches \cite{10.1145/3610542.3626152, yu2024instantstylegaussianefficientartstyle, bensadoun2024meta3dtexturegenfast} focus solely on altering the visual attributes of 3D models, for example, by generating textures and adjusting color properties. 
Given the interdependence of geometry and appearance, other works \cite{Haque2023InstructNeRF2NeRFE3, 10655916, 10376834} modify both local geometry and color attributes concurrently.

Among mesh-based methods, the standout approach, Text2Mesh \cite{t2m}, introduced by Michel \etal, leverages a neural network-based style field to modify the position and color of mesh vertices. 
However, the network's parameters are optimized using the CLIP \cite{Radford2021LearningTV} similarity loss between rendered images and text. 
Since CLIP does not directly generate images, the resulting model often lacks clear, semantically segmented textures, instead producing repetitive patterns across the entire surface, significantly differing from methods incorporating image generation models. 
Later, Dong \etal proposed Coin3D \cite{dong2024coin3d}, an advancement built on a multi-view generative model \cite{liu2023syncdreamer} combined with a geometric proxy. 
This approach employs a volume-controlled multi-view diffusion model, where the volume is obtained by sampling the mesh, to generate multiple views. 
The 3D model is then reconstructed from these views using NeRF \cite{mildenhall2020nerf} or NeuS \cite{wang2021neus} before re-extracting the mesh. 
However, it discards the initial mesh proxy, leaving the final output entirely dependent on the reconstructed model. 
Due to the limited 3D consistency of multi-view-generated images, this can result in overly blurred textures or shape misalignment, undermining geometric fidelity to the original proxy. 
Our proposed method retains both the neural style field and geometric priors while leveraging the high-quality textures provided by the diffusion model.

\section{APPROACH}
% \paragraph{Formulation}
% We target the generation of 3D building models with detailed appearance conditioned by a footprint, a hint of roof structure, and a style description. We input the footprint and roof structure as images, namely the mask and height sketch, and the style description as text. To this end, we propose GeoTexBuild, a modular framework that allows easy and accurate control and produces stylized 3D models in various building patterns.

We introduce GeoTexBuild, a modular framework that allows easy, accurate control and produces stylized 3D models in various building patterns. In this section, we briefly highlight key techniques closely related to our method in Sec.\ref{ssec:ExModule}, then give an overview in Sec.\ref{sssec:OV}, and details of each component in Sec.\ref{sssec:HMG} to \ref{sssec:AS}.

\subsection{Preliminaries}
\label{ssec:ExModule}
\subsubsection{ControlNet}
ControlNet \cite{zhang2023adding} is a powerful neural network architecture to guide text-to-image diffusion models \cite{10.5555/3495724.3496298} with spatially localized, task-specific image conditions. It attaches an additional trainable copy of the diffusion model to the backbone to inject the spatial information of the desired condition.
First, the spatial condition \(c_i\) is encoded by a tiny network, as \(c_f = E(c_i)\). Then, trainable clones with new parameters \(\theta_c\) are added to the frozen pre-trained U-net. The clones take the external conditioning vector \(c_f\) as input and are connected with zero convolutions \(Z(\cdot; \cdot)\). For a single neural block \(F(\theta)\), ControlNet outputs
\begin{equation}
    y_c = F(x; \theta) + Z(F(x + Z(c_f; \theta_{z1}); \theta_c); \theta_{z2}).
\end{equation}
For the entire diffusion model, multiple ControlNet blocks are applied to each encoder level to achieve fine control of different scales. It is a computationally efficient architecture, enabling high-speed, low memory consumption single-GPU training.

ControlNet enables multiple control methods, including edges and masks, while supporting both tandem and parallel conditioning over shape and color. This allows precise control of the footprint and detailed roof structures from low-cost freehand sketches, eliminating the need for 3D data.

\subsubsection{Text2Mesh}
Text2Mesh \cite{t2m} is a method for stylizing a given coarse mesh through natural language. It considers the given 3D mesh as an overall shape, and vertex attributes like color and displacement as style. There are three main components: a neural style field (NSF), a differentiable render, and a similarity comparator.
The input mesh \(\mathcal{M}\) is first normalized to a unit bounding box and fixed throughout the process. The neural style field has three MLPs: \(N_s\), \(N_c\), and \(N_d\), which map a point \(p\) in the unit space to a vector and a scalar, as
\begin{equation}
    (c_p, d_p) = NSF(\gamma(p)) \in (\mathbb{R}^3, \mathbb{R}),
\end{equation}
where \(\gamma(\cdot)\) is positional encoding. In each optimization step, the NSF is queried with vertices' coordinates \(\mathcal{V}\) to modify their attributes-color \(c_p\) and displacement \(d_p \cdot \vec{n_p}\) along the surface normal. \(n_\theta\) views of the modified mesh are rendered with multiple 2D augmentations. Then, semantic loss is computed by a pre-trained CLIP model \cite{Radford2021LearningTV}, as: \(L_{sim} = -\sum_S sim(S, \phi_t)\), where \(S\) and \(\phi_t\) are embeddings of renderings and text, \(sim(\cdot, \cdot)\) is the cosine similarity. Minimizing the loss means maximizing the similarity between the renderings and the text.

In our framework, we leverage the NSF to modify the position and color of the vertices while maintaining overall geometric fidelity with the generated mesh.

\subsubsection{Coin3D}
Coin3D \cite{dong2024coin3d} is a 3D-aware controllable mesh stylization method using a volume-controlled multi-view diffusion model and NeuS reconstruction. It consists of four steps: sampling and rendering the mesh proxy, generating the style image, forwarding multi-view diffusion, and reconstructing the 3D model.
More specifically, given a coarse shape \(P\), \(N_p\) surface points \(\mathcal{P}\) are sampled from it, and a front-view image \(I_f\) is rendered. Then, ControlNet (soft-edge and depth) generates the style image \(I_s\) with \(I_f\) a style prompt \(t\). After that, the multi-view diffusion generator \(f\) predict \(N_v\) consistent images \(I_{mv}^i\) under predefined camera poses \(\mathbf{c}_p^i\) as follows:
\begin{equation}
    I_{mv}^i = f(\mathcal{P}, I_s, \mathbf{c}_p^i).
    \label{mvd}
\end{equation}
% Going into the diffusion network \(f\), it is trained with the following objective: \(\min_{\theta} \mathbb{E}_{\tau,I_{mv}^i,\epsilon^i}\|\epsilon^i-\epsilon_{\theta}(I_{mv}^{i,\tau},\tau,c(I_s, \mathcal{F}^\tau, \mathbf{c}_p^i))\|\), where \(\theta\) is the diffusion network parameters, \(\epsilon\) and \(\epsilon_\theta\) are the noise added to image and predicted by the network, \(\tau\) is timestamp, \(c(\cdot)\) is the condition embedding, \(\mathcal{F}\) is the 3D control volume from surface points \(\mathcal{P}\).
After obtaining the multi-view images, they are used to reconstruct 3D shape following NeuS and volume-SDS. The result mesh is also extracted from NeuS.

In our framework, we use the volume-controlled multi-view diffusion as our stylized building image generator backbone.

\subsection{GeoTexBuild}
\subsubsection{Overview}
\label{sssec:OV}
Our framework, shown in Fig.\ref{fig:pipeline}, is built on key insights: \textbf{(A) the importance of footprints and roofs}, and \textbf{(B) the fact that geometry and appearance can be decoupled}. We therefore separate the generation of the roof and the appearance stylization into standalone stages, respectively.
We start with a given footprint and a height sketch. Stage 1, \textbf{height map generation}, outputs a regular and smooth height map; Stage 2, \textbf{geometry reconstruction}, outputs low-polygon and clean geometric models; and Stage 3, \textbf{appearance stylization}, enhances the mesh with detailed color and local geometries. The result mesh is exported in OBJ format for easy downstream applications or further manual modification.
This decomposition reduces the data dimensionality and complexity at each stage, which in turn lowers training costs and simplifies the process of editing or extracting intermediate results.

\subsubsection{Height Map Generation}
\label{sssec:HMG}
Our height map generator is based on a variant \cite{ghoskno_color_canny_controlnet} of ControlNet, which transfers a rectangular color palette to a regular colored image. To be suitable for building generation, we have the following considerations: (a) the grayscale height sketch serves as the primary control, where brighter areas indicate greater elevation. It can be quickly drawn by hand in seconds, and since the height is relative, the result can be shifted or scaled in later operations. (b) Freehand sketches may extend beyond the footprints; we input the footprints as masks \(I_m\) to process the sketches and achieve a stronger control with less confusion. (c) A building design may require a switch between different roofs; we design prompts for several fixed roof types and use different palette grid sizes for ControlNets.

\paragraph{Preprocessing} We first process the \(w_s \times w_s\) sketch to a square brightness palette with a grid size \(n_g\): 
\begin{equation}
\label{eq:preprosess}
    I_p^{i,j} = \frac{1}{w_p^2} \sum_{\substack{(i-1)w_p<x \leq iw_p \\ (j-1)w_p<y \leq jw_p}} I_s(x,y),
\end{equation}
where  \(w_p = w_s / n_g\) is the width of the palette grid, \(I_p^{i,j}\) is the pixel value of the square in the i-th row and j-th column. We use a default size of \(w_s = 512\) in ControlNet. The brightness palette is then masked by the footprint to create a clear boundary, as:
\begin{equation}
    I_{mp} = I_p \odot I_m,
\end{equation}
where \(\odot\) is the Hadamard product. Preprocessing with an adjustable grid size helps balance the difficulty and precision of user input, since users sketch with solid color blocks rather than gradients.

\paragraph{Applying ControlNet} To assist ControlNet in understanding different possible types of roofs from a single brightness palette, we construct several fixed patterns of prompts \(t\), describing whether the roof has multiple parts and its general shape. The prompts are listed in Tab.\ref{tab:roof-prompts}. Then, the generating process can be represented as: 
\begin{equation}
    I_h = CN(I_{mp}, t).
\end{equation}
To learn the map between \(I_{mp}\) and \(I_h\), we trained the ControlNet with a dataset of roof reconstructions of real-life architectures - Building3D \cite{Wang2023Building3DAU}. See details explained in Sec.\ref{ssec:TrainCN} and supplementary material.

\begin{table}
\centering
\caption{Prompts for Different Roof Types}
\label{tab:roof-prompts}
\resizebox{1.0\linewidth}{!}{
\begin{tabular}{ccccc}
\toprule
No. & Connectivity & Node Degree & Z-Coord. Range & Prompt \\ \midrule
1 & Multi-Piece           & \(>4\)(complex)          & Pitched                     & Grayscale depth-map, multiple parts of complex slopes and layers, in polygon shape, black background \\
2 & Multi-Piece           & =3, 4(medium)            & Pitched                       & Grayscale depth map, multiple parts of slopes and layers, in polygon shape, black background \\
3 & Multi-Piece           & \(<3\)(simple)           & Pitched                     & Grayscale depth-map, multiple simple shed layers, in polygon shape, black background \\
4 & Multi-Piece           & /                        & Flat                        & Grayscale depth-map, multiple flat layers, in polygon shape, black background \\
5 & Single-Piece          & \(>4\)(complex)          & Pitched                      & Grayscale depth map, a joint part of complex combinations of slopes, in polygon shape, black background \\
6 & Single-Piece          & =3, 4(medium)            & Pitched                       & Grayscale depth map, a joint part of simple slopes, in polygon shape, black background \\ 
7 & Single-Piece          & \(<3\)(simple)           & Pitched                     & Grayscale depth map, a simple shed layer, in polygon shape, black background \\
8 & Single-Piece          & /                        & Flat                        & Grayscale depth map, a flat layer, in polygon shape, black background \\
\bottomrule
\end{tabular}}
\end{table}

\subsubsection{Geometry Reconstruction}
The height map goes through a series of operations to reconstruct the geometry of the building.

\paragraph{Image processing} As a diffusion model with zero convolution generates the relative height map, it is inevitable to have some burst noise near the edges of the controls. Other noise also exists in plain areas because of the nature of diffusion models. We filter the image with a bilateral filter \cite{10.5555/938978.939190} \(\mathcal{F}_b\) to smooth the height value while preserving the edges of different parts, with filter size set to 7, \(\sigma_{gr} = 9, \sigma_{gs} = 55\).
% \begin{equation}
%     I_{bf}(x) = \frac{1}{W(x)} \sum_{x' \in \Omega} {I_h(x') \cdot g_r(\|x - x'\|)\cdot g_s(\|x - x'\|)},
% \end{equation}
% where \(W(x)\) is the normalization factor, \(\Omega\) is neighborhood of x, \(g_r\) and \(g_s\) are Gaussian kernels. To ensure sufficient smoothing, the filter size is set to 7 with \(\sigma_{gr} = 9, \sigma_{gs} = 55\). 
We erode the mask by one pixel with a morphological operator \(\mathcal{E}(\cdot)\) to further remove bright spots near the edges, as:
\begin{equation}
    I_{bfe} = \mathcal{F}_b(I_h) \odot \mathcal{E}(I_m).
\end{equation}

\paragraph{Separating and polygon fitting} In \(I_{bfe}\), we fit each color zone to a polygon with Vtracer \cite{visioncortex_vtracer}, a raster-to-vector graphics converter. In this module, the pixels are first clustered based on differences in brightness, then hierarchically laid on the canvas. To prevent color zones from missing or holes, we set color precision to 6, meaning a zone has to have a color difference larger than \(256/6 \approx 42\) to be considered a separate cluster. Those clusters go through path walking, simplification, and curve fitting to be vectorized into polygons. We extract each polygon as an independent roof part from the output graph. With them, we obtain the position of the vertices in the exterior \(\{\partial \mathcal{S}_r\}\) of the reconstructed roof \(\{\mathcal{S}_r\}\). 

\paragraph{Lifting image to point cloud} We reconstruct a dense point cloud from the height map pixels encircled by each polygon. The positions and values of pixels in \(I_{bfe}\) are set directly as the \((x, y)\), and z-coordinates of the reconstructed points. We clean the point cloud by removing statistical outliers and those close to the ground (\(z < h_{min}\)). 

\paragraph{Reconstructing surfaces} Each part of the point cloud is sent to a normal estimator and a Poisson surface reconstructor \cite{10.5555/1281957.1281965} with octree depth 6, after which fairing is applied according to the roof type. The reconstructed roof surfaces \(\{\mathcal{S}_r\}\) remain open, as they originate from an open-boundary point cloud with consistently oriented normals. This helps us to refine the ratio of the roof's height and absolute building height via a scaling factor-building parameter \(p_b\) with \(h_{\mathcal{M}_b} = h_{facade} + h_{roof}\) and \(h_{roof} = h_{facade} \times p_b\). The building height and \(p_b\) can be chosen freely to meet architectural preferences. However, open surfaces restrict post-reconstruction trimming, which would disrupt the smooth boundary and lead to stripy facades. To address this, we take advantage of Boolean operations: We construct meshes \(\{\mathcal{M}_r\}\) of each roof part by extruding them down from the desired height. Meanwhile, prisms \(\{\mathcal{M}_{\partial \mathcal{S}_r}\}\) are extruded bottom-up from \(\{\partial \mathcal{S}_r\}\). Then, the final mesh is:
\begin{equation}
    \mathcal{M}_b = \cup( \mathcal{D}(\{\mathcal{M}_r\} \cap \{\mathcal{M}_{\partial \mathcal{S}_r}\})),
\end{equation}
where \(\cap\) refers to the intersection of corresponding meshes in two sets, and morphological operator \(\mathcal{D}(\cdot)\) refers to 1 unit dilation in \(x\), \(y\) directions for each object. The dilation compensates for the erosion applied before and eliminates holes or cracks caused by polygon fitting.

\begin{figure}
    \centering
    \includegraphics[width=1.0\linewidth]{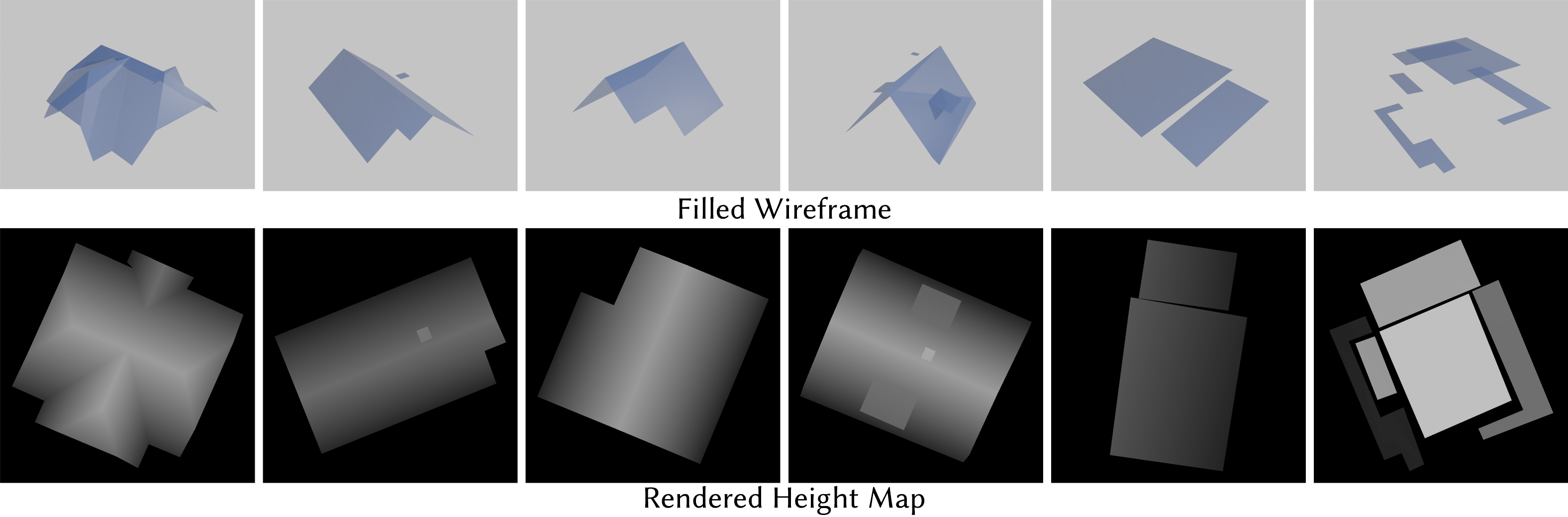}
    \caption{Random data samples from rendered height maps of Building3D dataset. The dataset offers over 32k building reconstructions. The samples belong to types 5, 2, 6, 1, 3, 4, in Tab.\ref{tab:roof-prompts}, respectively.}
    \label{fig:building3d}
    \Description{}
\end{figure}

\subsubsection{Appearance Stylization}
\label{sssec:AS}
To overcome both the repetitive-pattern texture problem of Text2Mesh \cite{t2m} and the misalignment geometry of Coin3D \cite{dong2024coin3d}, we propose a method to combine NSF and multi-view diffusion, but exclude CLIP similarity loss or NeuS reconstruction.

The outcome mesh \(\mathcal{M}_b\) from the former stage acts as the reference for NSF and the proxy for volume-controlled multi-view diffusion. After remeshing to densify triangles on surfaces, we have a sufficient resolution mesh for the NSF. Following the instructions of Coin3D, this mesh is sampled to produce surface points \(\mathcal{P}\), and rendered to produce a style image \(I_s\) with a user prompt \(t\).

The multi-view images \(I_{mv}\) are generated by Eq.\eqref{mvd}, and set to the ground truth image for the NSF after an alignment transformation \(\mathcal{T}_a(\cdot)\). By this, the NSF is no longer constrained by the CLIP model, allowing it to avoid repetitive textures with poor semantic coherence and instead focus on learning the textures generated by the multi-view diffusion model. Additionally, the affine transformation \(\mathcal{T}_a(\cdot)\) aligns the generated image to the reference mesh in a pre-rendering process. By operating directly on the reference mesh, NSF bypasses the NeuS reconstruction process, avoiding geometric misalignment that can arise when generated images lack multi-view consistency.

More precisely, in each NSF optimization step, we render \(N_v\) views \(I_{mv}\) and masks \(M_{nsf}\) of the NSF modified mesh from predefined multi-view cameras \(\mathbf{c}_p\). Image loss and mask loss are calculated from the rendered and the diffusion-generated images \(\mathcal{T}_a(I_{mv}), \mathcal{T}_a(M_{mv})\) as follows:
\begin{equation}
    L = \lambda_{rgb}L_{rgb} + \lambda_{ssim}L_{ssim} + \lambda_{lpips}L_{lpips} + \lambda_{mask}L_{mask},
\end{equation}
where \(L_{rgb}\) is \(L_1\) loss per pixel color, \(L_{mask}\) is \(L_2\) loss per mask pixel value (soft mask), and \(\lambda_{rgb} + \lambda_{ssim} + \lambda_{lpips} = 1\) (\(\lambda_{mask}\) not restricted because \(L_{mask}\) do not contribute to color values). By adding mask loss \(L_{mask}\), we have sufficient regularization for the overall geometry. To create better details, we modified the displacement output of the NSF to a vector \((d_x, d_y, d_z)\) that directly adds to the vertices' coordinates and is no longer restricted to surface normal directions.

\section{EXPERIMENTS}
To fully complete the framework, we customized the ControlNet model on depth renderings of the Building3D \cite{Wang2023Building3DAU} dataset for height map generation (Sec.\ref{ssec:TrainCN}), built the NSF+MV-diffusion for appearance stylization (Sec.\ref{ssec:NSFMV}), and evaluated the complete framework (Sec.\ref{ssec:ResFramework}). The ablation studies on geometry reconstruction are in the supplementary material.
% We conducted ablation studies on geometry reconstruction (Sec.\ref{ssec:Stage2Abl}), and evaluated the complete framework (Sec.\ref{ssec:ResFramework}).

\subsection{Height Map Generation}
\label{ssec:TrainCN}
\subsubsection{Customizing ControlNet}
Our customized ControlNet learns the mapping from masked brightness palettes to a height map. We prepared paired data of depth renderings from the dataset Building3D (Fig.\ref{fig:building3d}), classified them, and designed prompts for several roof types (Tab.\ref{tab:roof-prompts}). See detailed descriptions in the supplementary material.

\subsubsection{Comparing different control combinations}
\begin{figure}
    \centering
    \includegraphics[width=\linewidth]{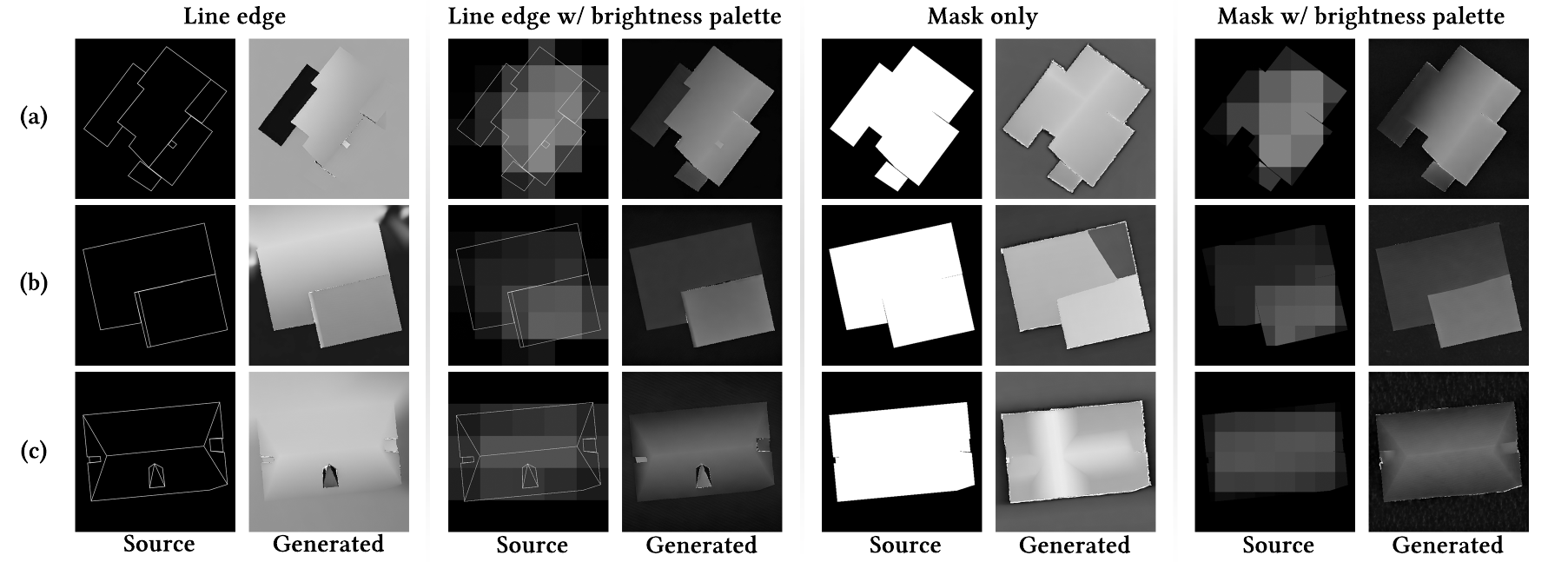}
    \caption{Four customized ControlNet models that input different geometric controls are trained with the same hyperparameters. The corresponding prompts used when inference is No.6 for (a)(b), and No.5 for (c). (Prompt index in Tab.\ref{tab:roof-prompts})}
    \Description{}
    \label{fig:different_control}
\end{figure}

We compare different control combinations of the generation process here to demonstrate the superiority brought by the masked brightness palette. Here are four models trained with the same hyperparameters in Fig.\ref{fig:different_control}, but input different controls: (1)line edges only, (2)line edges and brightness palette, (3)masks only, (4)masks and brightness palette. The result shows that, first, in (1)(3), the height is generated randomly without the brightness palette; second, in (1), the line edges cause ambiguity in distinguishing the internal and external. As for (2), drawing line edges and height sketches simultaneously needs extra consideration of their compatibility, but shows similar quality compared to (4). Although (4) has lost some delicate details, it generates less noise at the edges of the inner parts, which reduces the difficulty of filtering.

\begin{figure}
    \centering
    \includegraphics[width=1\linewidth]{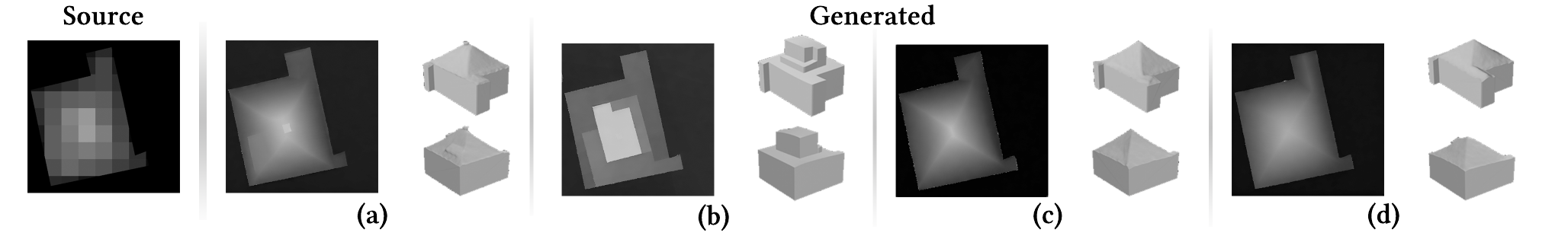}
    \caption{By changing different prompts when inputting the same control, we demonstrate that the model can understand different possible types of roofs from a single brightness palette. The corresponding prompts are No.1 for (a), No.4 for (b), No.5 for (c), No.6 for (d). The prompt in (a) indicates more layers, such as chimneys and dormer windows, while similar prompts (c) and (d) produce similar results.}
    \Description{}
    \label{fig:different_prompt}
\end{figure}

Additionally, we demonstrate that by altering prompts for roof types in Tab.\ref{tab:roof-prompts}, the model can produce corresponding height maps as in Fig.\ref{fig:different_prompt}.

\subsection{NSF + MV Diffusion Stylization}
\label{ssec:NSFMV}
Our stylization module enhances the mesh with detailed geometry and semantically clear textures, while preserving the overall structure and the footprint edges inherited from the first two stages. We compare our stylization method with the outstanding geometry-controlled generative methods: Text2Mesh \cite{t2m} and Coin3D \cite{dong2024coin3d}, with the same geometric conditioning and style prompts in this section.

\begin{figure}
    \centering
    \includegraphics[width=\linewidth]{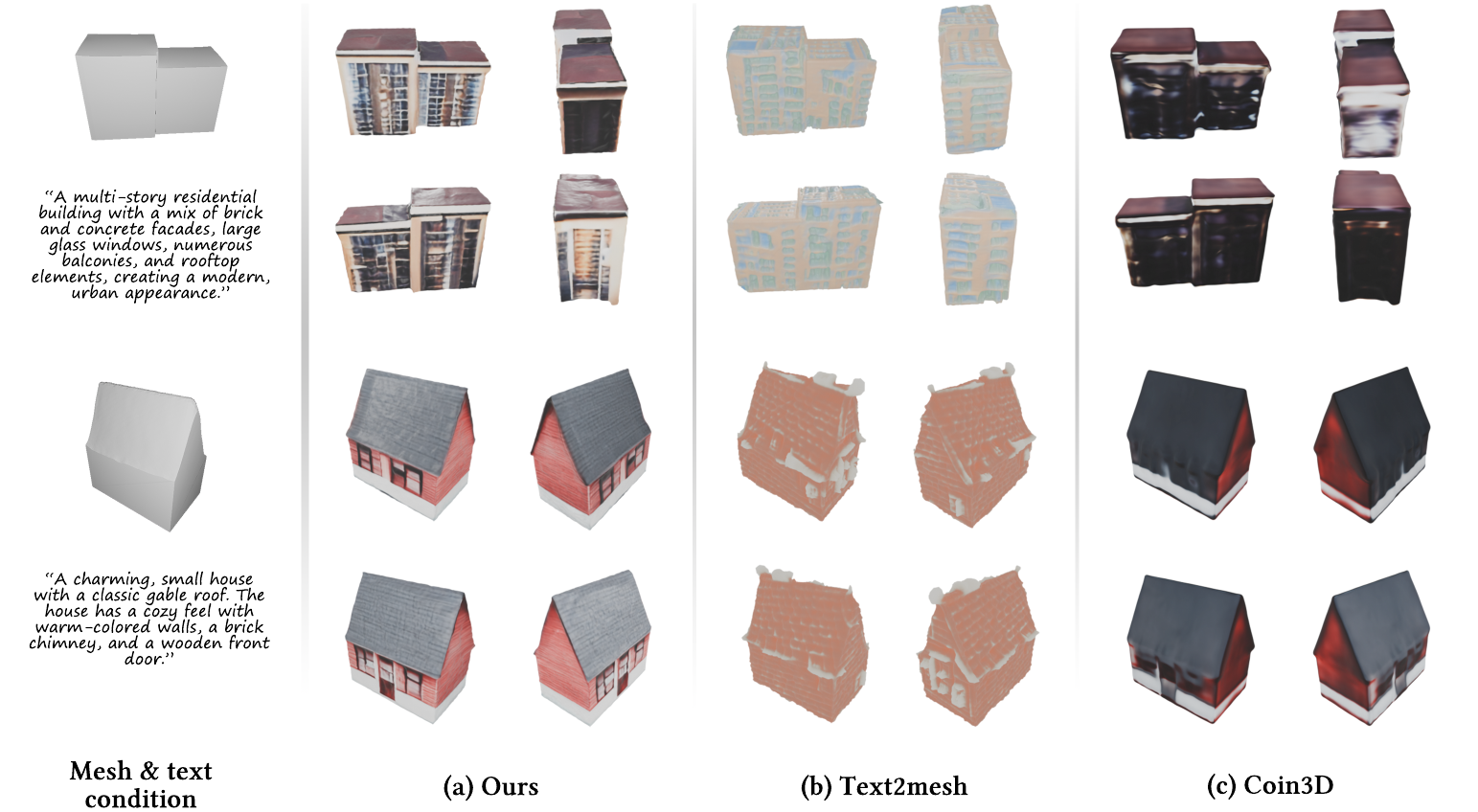}
    \caption{We compare our stylized textured mesh with geometry-controlled stylization methods Text2Mesh and Coin3D, the images are rendered in Blender with four fixed camera poses.}
    \Description{}
    \label{fig:compare_t2m_c3d}
\end{figure}

\begin{figure}
    \centering
    \includegraphics[width=\linewidth]{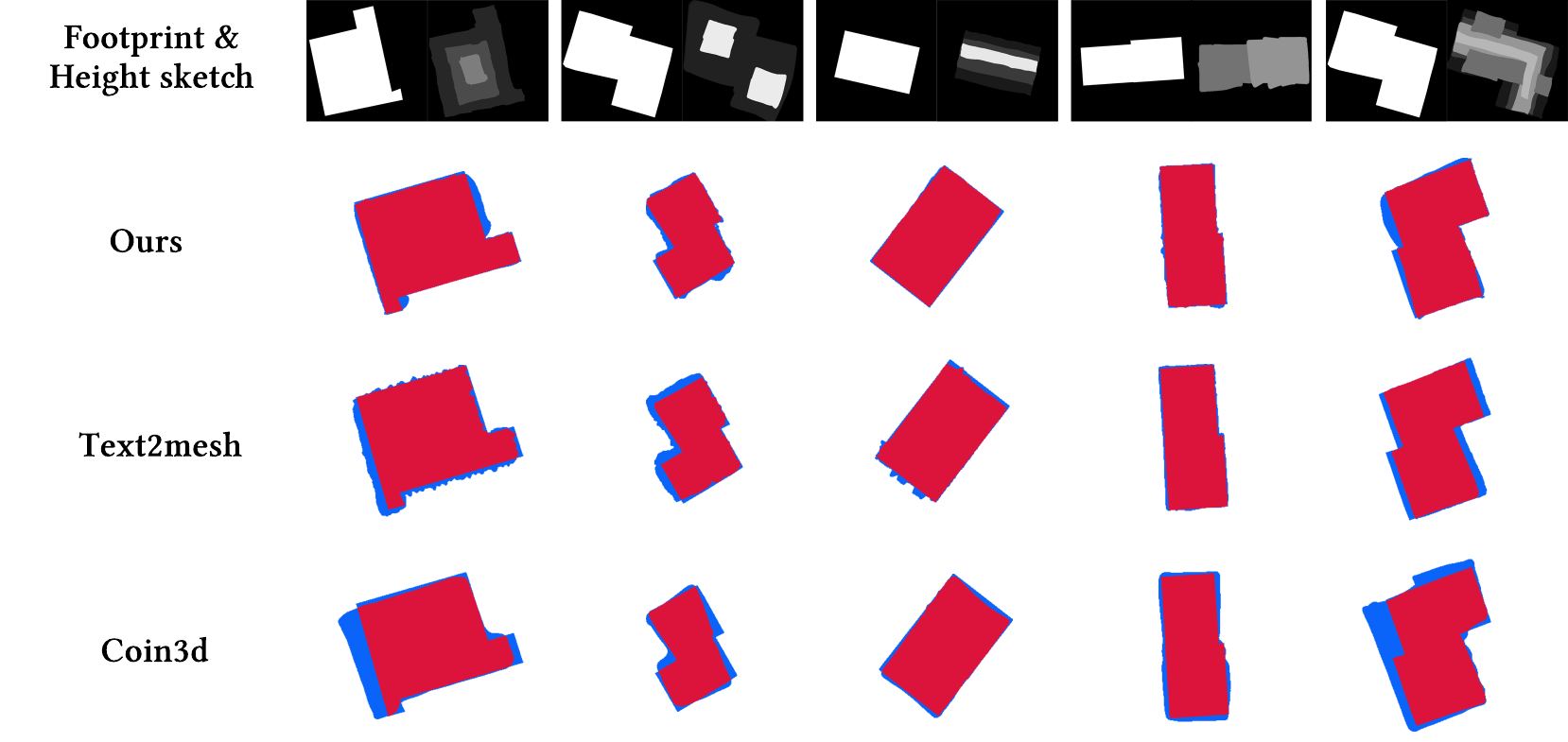}
    \caption{We compare the ability to preserve the footprints by visualizing the IoU of the initial footprints and final mesh top-down projections. The overlaps are displayed in Crimson, other parts are in cobalt blue. The corresponding footprints and height sketches are also displayed in the upper row.}
    \Description{}
    \label{fig:iou_vis}
\end{figure}

\paragraph{Qualitative Comparison} We show the renderings of the final extracted mesh in Fig.\ref{fig:compare_t2m_c3d}. The IoU visualization of the initial footprints and final mesh top-down projections is in Fig.\ref{fig:iou_vis}. In the rendered images, Text2Mesh exhibits repetitive textures across the entire mesh surface, often inaccurately placing windows on the roof (top row) and failing to differentiate between the roof and facade walls (bottom row). Additionally, it sometimes struggles to produce clear colors. Coin3D, in contrast, suffers from significant texture degradation in the NeuS reconstruction due to the limited 3D consistency of the generated images. Inaccuracies in opaque density estimation during neural inverse rendering result in the blending of front and back surface colors. Our method leverages the strengths of both approaches, achieving high-quality results. In the IoU visualization, Text2Mesh generates uneven surfaces, leading to a reduction in its IoU. Coin3D experiences a significant loss as it discards the mesh proxy and relies solely on reconstruction. Our method, by contrast, achieves strong performance.

\paragraph{Quantitative Comparison} We present the CLIP score \cite{Radford2021LearningTV}, ImageReward \cite{xu2023imagereward}, IoU, and execution time in Tab.\ref{tab:metrics} to assess text-image alignment, the perceptual quality of the renderings, control quality, and efficiency. Our method achieves the overall best metrics, which emphasize the effectiveness of optimizing NSF with multi-view generated images. Note that Text2Mesh achieves a high CLIP score because it is the primary optimization objective. We achieved the shortest execution time among the methods by leveraging MV images, which reduce the iterations required to optimize the NSF. Besides, since 3D shape reconstruction using NeuS is no longer needed compared to Coin3D, additional computational overhead is eliminated.

\begin{table}
  \centering
  \caption{We report quantitative evaluation of visual quality, footprint control quality, and execution time. Execution time of ours is in S1+S2+S3. All models were tested on an RTX3090, except TripoSR and Hunyuan3D from online services: https://www.tripo3d.ai/, https://3d.hunyuan.tencent.com/.}
  \resizebox{1.0\linewidth}{!}{
    \begin{tabular}{lcccccc}
    \toprule
    Gen. Type & \multicolumn{3}{c}{Geo-text Conditioned} & Text Conditioned & \multicolumn{2}{c}{Image Conditioned} \\
    \cmidrule(lr){2-4}\cmidrule(lr){5-5}\cmidrule(lr){6-7} Method & Ours & Text2Mesh &  Coin3D & MVDream+3DGS & TripoSR 2.5 & Hunyuan3D 2.0 \\
    \midrule
    CLIP Score $\uparrow$ & \cellcolor{orange!40}0.2537 & \cellcolor{yellow!40}0.2428 & 0.2208 & 0.2199 & 0.1912 & 0.2069  \\
    ImageReward $\uparrow$ & \cellcolor{orange!40}-0.2940 & -1.5151 & \cellcolor{yellow!40}-1.3479 & -1.8994 & -2.2700 & -2.2643  \\
    IoU $\uparrow$  & \cellcolor{orange!40}0.9083 & \cellcolor{yellow!40}0.8763 & 0.7790 & - & - & - \\
    Execution Time $\downarrow$ & 12s+9s+$\sim$7min & $\sim$19min & $\sim$28min & $\sim$22min & $\sim$260s & $\sim$280s \\
    \bottomrule
    \end{tabular}}
  \label{tab:metrics}%
\end{table}%

\subsection{Results of the Entire Framework}
\label{ssec:ResFramework}
In this section, we compare our framework with other generative pipelines that use text or image conditions to demonstrate the efficiency and importance of geometry controls in building generation.

\begin{figure}
    \centering
    \includegraphics[width=\linewidth]{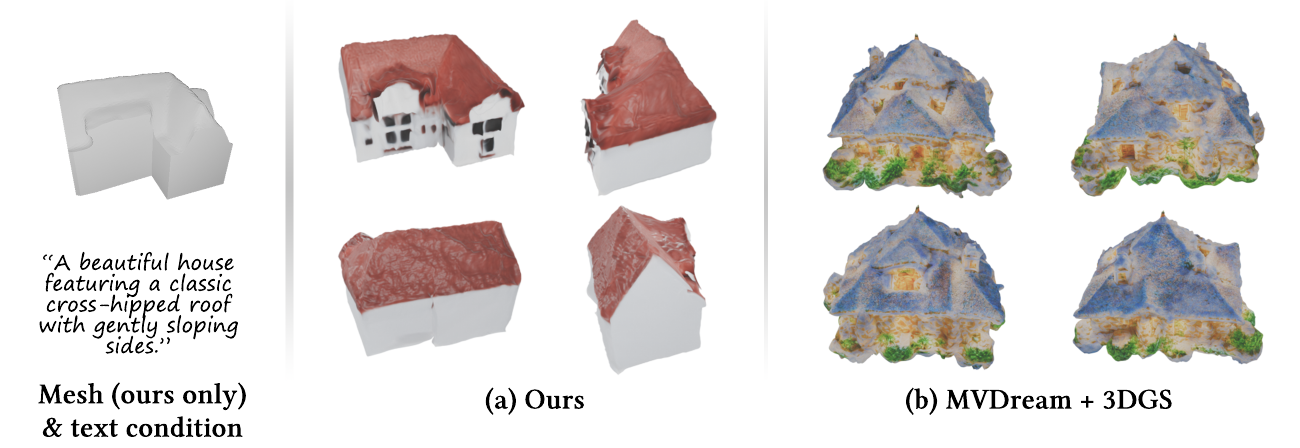}
    \caption{We compare our framework with the text-to-3D method MVDream+3DGS to highlight the importance of geometry control over relying solely on text descriptions.}
    \Description{The mesh of 3DGS is extracted using the method described in DreamGaussian.}
    \label{fig:compare_textgen}
\end{figure}

\paragraph{Comparison with text-to-3D} We compare our framework with MVDream+3DGS pipeline \cite{shi2023MVDream, kerbl3Dgaussians, ukarapol2024gradeadreamer} in Fig.\ref{fig:compare_textgen}. We show the generated mesh from stage 2 and the text prompt in the first column as input to our stylization module. The same text prompts are used as input to MVDream. Since MVDream relies solely on a text prompt as input, it generates shapes freely without exhibiting any control over the geometry structure. Additionally, it places vegetation around the building, complicating the extraction of a flat exterior wall surface. The CLIP score and ImageReward are also shown in Tab.\ref{tab:metrics}.

\begin{figure}
    \centering
    \includegraphics[width=\linewidth]{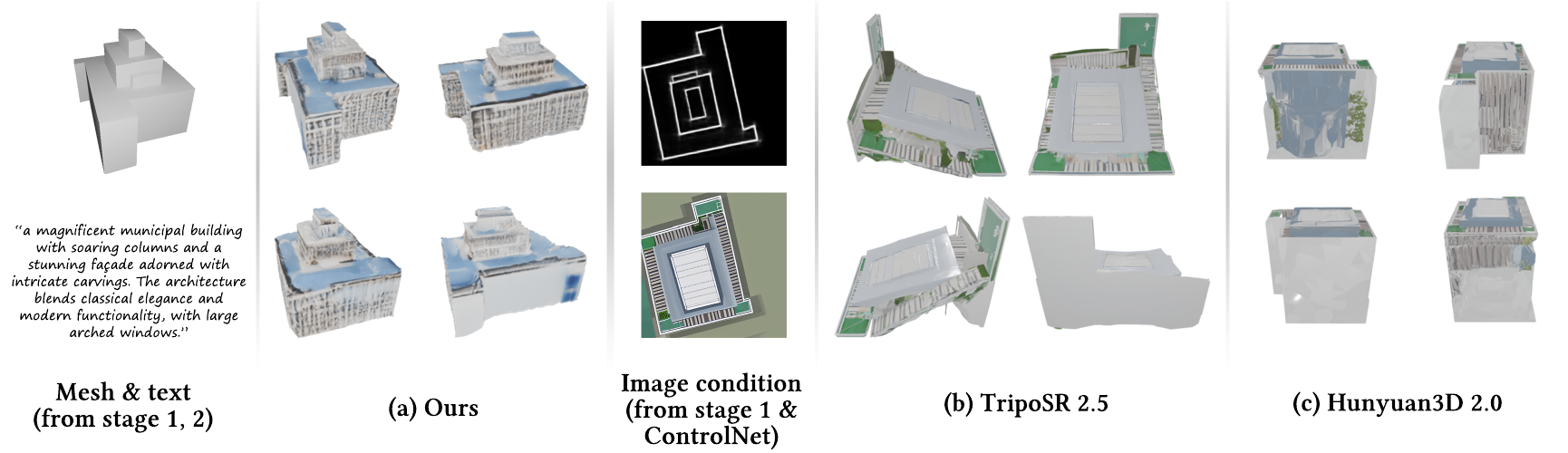}
    \caption{We compare our framework with industrial-level image-to-3D models, demonstrating that despite their strong geometric understanding, they struggle to generate reasonable building models from a single image. In contrast, modules across each stage work together in our framework and effectively achieve the goal.}
    \Description{}
    \label{fig:compare_imagegen}
\end{figure}

\paragraph{Comparison with image-to-3D} We compare our framework with TripoSR \cite{tochilkin2024triposrfast3dobject} and Hunyuan3D \cite{zhao2025hunyuan3d20scalingdiffusion}, two large-scale, production-ready generative models, in Fig.\ref{fig:compare_imagegen}. These models demonstrate strong geometric understanding and texture generation capabilities, producing high-quality 3D assets when given appropriate geometric guidance. However, they rely on a single condition—either image or text—limiting control over the generation process. For a fair comparison, we provide these models with building roof design images generated by ControlNet using the soft edge of our stage 1 result and text prompts as conditions. Despite this, they still struggle to interpret the meaning of the image condition accurately, though they demonstrate some control over the shape of the footprints. The CLIP score and ImageReward of their generated models are also shown in Tab.\ref{tab:metrics}.

\section{CONCLUSION}
We presented GeoTexBuild, a modular framework for generating textured 3D building models from a footprint, height sketch, and style prompt. Its design draws upon critical observations in city planning and architectural analysis, combined with foundational principles from computer graphics. Three components achieved height map generation, geometry reconstruction, appearance stylization, and built up the whole system.

We accomplished fine-grain control of footprint area and roof structures by customizing ControlNet. Then, we achieved detailed mesh stylization by proposing an NSF + MV diffusion method. This method eliminated both the repetitive-pattern texture and misalignment geometry problems when stylizing building meshes. Experiments and comparisons with geometry-controlled stylization and text/image conditioned 3D generation methods proved the effectiveness of our framework and each module.

GeoTexBuild is built for architects and city planners, for quickly transitioning conceptual designs to downstream tasks. It bridges creative vision and technical execution, enabling highly customized designs. It lays the groundwork for innovations like generative design integration and automated urban aesthetics analytics.

\paragraph{Limitations and Further Research} As discussed in Section \ref{ssec:ResFramework}, large 3D generative models \cite{tochilkin2024triposrfast3dobject, zhao2025hunyuan3d20scalingdiffusion}, trained on comprehensive 3D datasets with extensive GPU resources, may yield superior fine geometry and texture resolution. However, constraints related to data formats and computational capacity have precluded their integration into our pipeline. Investigating methods to integrate geometric control beyond renderings or text, while reducing the training costs of large 3D generative models to facilitate domain-specific fine-tuning, represents a promising avenue for future research.

\bibliographystyle{ACM-Reference-Format}
\bibliography{ref}

% \clearpage
\vspace{\baselineskip}

\appendix

{\noindent \Large \bf Supplementary Material}

\section{ControlNet Training Details}

\subsection{Dataset}
Our customized ControlNet \cite{zhang2023adding} learns the mapping from masked brightness palettes to a height map; thus, we need to create corresponding paired data. Open-source maps and city data, such as OpenStreetMap \cite{openstreetmap}, often do not contain building height data of sufficient accuracy to show the height variation inside a house's roof. We turned to city-scale building reconstruction data and extracted height information from their roofs. Building3D \cite{Wang2023Building3DAU} offers over 32k building reconstructions in multiple data forms, including point cloud, wireframe, and mesh. We chose wireframes because it is accurate enough without being as noisy as naive point clouds. We inserted surfaces in the wireframes and rendered orthogonal top views in Blender; the depth maps and masks were used as paired data in ControlNet training. We augmented the dataset \(8 \times\) by rotation, flipping, and random cropping. 

\paragraph{Prompts for roof types} We classified the roofs based on the top structure and geometry of the wireframes. The wireframes were regarded as graphs, and divided into 8 types according to the Connectivity, Node Degree, Z-Coordinate Range columns in Tab.\ref{tab:roof-prompts}. We simply describe these typical features in the prompt words without incorporating quantitative details, leaving room for more effective methods of classification and prompt construction.

\subsection{Training}
The rendered depth maps were used both as inputs and as ground truths. As inputs, they went through the preprocessing described in Sec. \ref{sssec:HMG}, which outputs a masked brightness palette. We set \(n_g\) a random integer from 5 to 9 to enable conditioning in different precisions. As ground truths, the depth maps are used to compute the noise predictor loss in the diffusion process.

We trained the ControlNet from the Stable Diffusion 1.5 \cite{stable_diffusion_1_5} base model on 260k paired images up to 32618 global steps. The batch size was set to 4 with gradient accumulation steps set to 4. The SD decoder was unlocked during the last 1000 steps with a learning rate change from 1e-5 to 2e-6, following the instructions of the ControlNet author. The entire training process was completed with a single Nvidia RTX 3090 GPU in 62 hours.

\paragraph{This ControlNet does more than image reconstruction} We use rendered depth as both inputs and GT, so it seems like we are just doing image reconstruction. Recap that we target generating from hand-drawn sketches, but the buildings from the real world do not have paired height sketches. Thanks to the preprocessing (Eq.\ref{eq:preprosess}) that fills the gap between the rendered depth and the hand-drawn sketches, we can train the model in a reconstruction way, but perform inference in a generation way. Also, at inference time, the palette grid size can be adjusted: a larger \(n_g\) (smaller grid cells) improves control when precision is low (the generated height map does not match the sketch), while a smaller 
\(n_g\) (larger grid cells) adds flexibility when the sketch is ambiguous (or the generator mechanically matches the sketch). In practice, we found that the generator performs well, even with \(n_g = 24\), which is much larger than in training (5—9), enabling highly precise control. Conceptually, preprocessing acts like a multi-scale pyramid, by average pooling, to help the model interpret varying levels of control.

\section{Geometry Reconstruction Details}
\label{sec:Stage2Abl}
Stage 2 contains image and geometry operations without any trainable neural blocks. Although it is labeled “reconstruction”, it does not reconstruct meshes from the Building3D dataset. Instead, it aims to add surfaces and assigns height (via \(p_b\)) to "reconstruct" the 2D output of S1 into a solid model. We built the module with Easy3D \cite{Nan2021} and Blender. 

\paragraph{Range of Shape Reconstruction}
\label{para:RSR}
In practice, our method imposes minimal constraints on generated shapes. Special shapes, for example, Tower-like buildings, are treated as having negligible extruded height (\(h_{facade} \to 0, p_b \to \infty\), and \(h_{\mathcal{M}_b} \approx h_{roof}\)). When the roof shapes vary along with the input sketches, all convex shapes are generable.

\subsection{Parameter Settings}
We prioritize computational efficiency while ensuring sufficient accuracy. These parameters typically have limited dynamic ranges (e.g., octree depth: 6-9; color precision: 4-8; filter size: 3, 5, 7, 9). Through empirical testing, we observed clear performance trends and selected the minimal values that still yield reliable results across all test data.

\subsection{Ablation Studies}
Here, we demonstrate the results of the ablation studies.

\begin{figure*}
    \centering
    \includegraphics[width=1\linewidth]{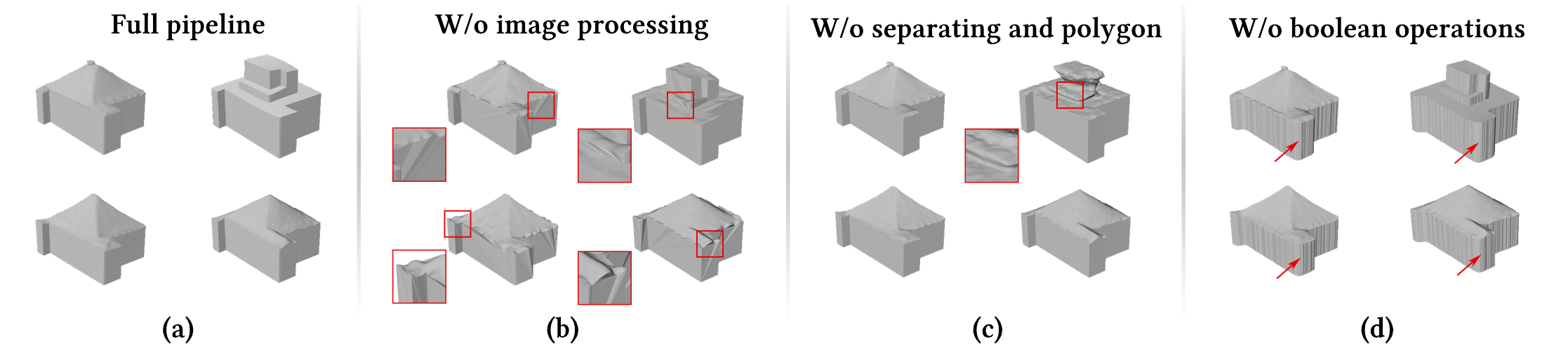}
    \caption{Results of ablation studies for geometry reconstruction. Please note the irregular shapes and uneven edges in the red boxes or pointed out by arrows, which are direct consequences of omitting crucial steps.}
    \Description{}
    \label{fig:geometry_recon_ab}
\end{figure*}

\paragraph{w/o image processing} The reconstructed surfaces are shown in the (b) part of Fig.\ref{fig:geometry_recon_ab}. The image processing helps to remove noise and keep a sharp boundary of the reconstructed roof parts.

\paragraph{w/o separating and polygon fitting} The reconstructed surfaces are shown in the (c) part of Fig.\ref{fig:geometry_recon_ab}. In the absence of the separation and polygon fitting, various roof parts merge, resulting in over-smooth surfaces or bread-like boundary artifacts in Poisson reconstruction. Although the impact on a single-layer roof is relatively minor, it becomes catastrophic when applied to multi-layer roofs.

\paragraph{w/o boolean operations} The reconstructed surfaces are shown in the (d) part of Fig.\ref{fig:geometry_recon_ab}. Replacing the boolean operations with simple trimming will destroy the smooth boundary we try to preserve, leading to stripy facades.

\section{NSF + MV Diffusion Details}
\label{asec:nsf}
\subsection{NSF Details}
\paragraph{Necessity of the NSF}
As shown in Fig.\ref{fig:compare_textgen} and \ref{fig:alignment}, MV-diffusion may produce slight shape variations in small regions such as eaves or dormers. NSF displacement bridges the gap between the S2 shape prior and the geometry implied by MV-diffusion. Without it, directly applying color can cause misalignment and visual artifacts.

\paragraph{Hyperparameter Settings}
In our experiments, we observed that using the default settings in Text2Mesh \cite{t2m} results in an over-deep network for multi-view stylization. Due to spectrum bias in deep neural networks, high-frequency details are often lost, necessitating the use of a Fourier Feature Transform layer \cite{10.5555/3495724.3496356} at the network's input to mitigate it. However, applying a high \(\sigma\) in the Fourier Feature Transform to deep NSFs can degrade color learning, causing the network to capture only approximate hue and brightness rather than the true color. This issue can be alleviated by using shallower networks, lower \(\sigma\), and increasing the number of Fourier feature channels. Nevertheless, a balance between high-frequency content and accurate color representation must be maintained. To achieve this, we retain a high \(\sigma = 6\) but compensate by reducing the network depth (depth of \(N_s, N_c, N_d\) set to 1, 2, 2) and expanding the Fourier feature width (4096) and network width (512). We optimize the NSF with a full-batch training over 220 steps.

\subsection{Alignment Transformation}
\label{assec:align}
\begin{figure*}
    \centering
    \includegraphics[width=\linewidth]{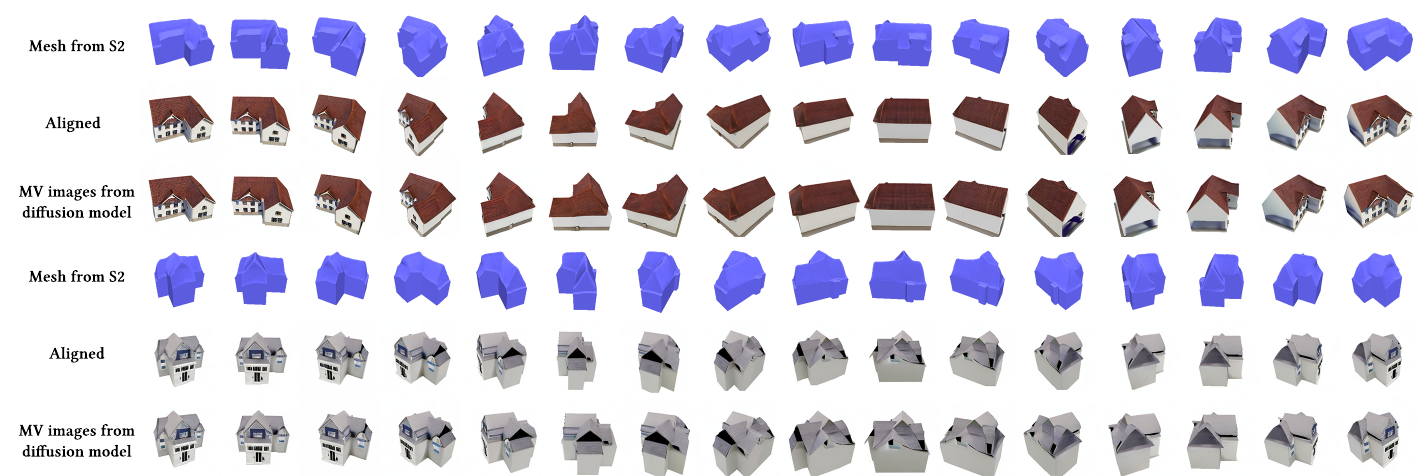}
    \caption{Renderings of mesh from stage 2, generated images from the multi-view diffusion, and aligned images. Notice the large displacement of the generated image, especially from the 6th to 12th views. The first and second buildings correspond to results in Fig.\ref{fig:fop2} and Fig.\ref{fig:fop3}.}
    \Description{}
    \label{fig:alignment}
\end{figure*}

We illustrate the raw image from the MV diffusion and the renderings of meshes from stage 2 in Fig.\ref{fig:alignment}. Despite applying volume-controlled MV diffusion, significant content distortion persists in the intermediate view, leading to geometry misalignment with the mesh proxy. To preserve the mesh reference within the NSF, we apply an affine transformation to the generated image, ensuring alignment with the bounding box of the mesh renderings. However, any content extending beyond the canvas remains clipped and cannot be recovered. 

To apply the Alignment, paired points of the source image (from render) and the destination image (from MV diffusion) are first selected: we choose the top-left, top-right, and bottom-left corners of the two bounding boxes as \(p_1, p_2, p_3\). We solve for the transformation matrix A that maps the source points to the destination points:
\begin{equation}
    \begin{bmatrix}
    x_{dst}^{i} \\
    y_{dst}^{i}
    \end{bmatrix}
    =
    A
    \begin{bmatrix}
    x_{src}^{i} \\
    y_{src}^{i} \\
    1
    \end{bmatrix}.
\end{equation}
Then, we apply this transformation to obtain the locations of pixels in the aligned image, as
\begin{equation}
\begin{split}
    \begin{bmatrix}
    x_{dst} \\
    y_{dst}
    \end{bmatrix}
    &= A^{-1}
    \begin{bmatrix}
    x \\
    y \\
    1
    \end{bmatrix}, \\
    I_{aligned}(x,y) &= I_{dst}(x_{dst}, y_{dst}) \text{ (with interpolation)}.
\end{split}
\end{equation}
We use \(I_{aligned}\) as ground truths of the NSF training.

\begin{figure*}
    \centering
    \includegraphics[width=\linewidth]{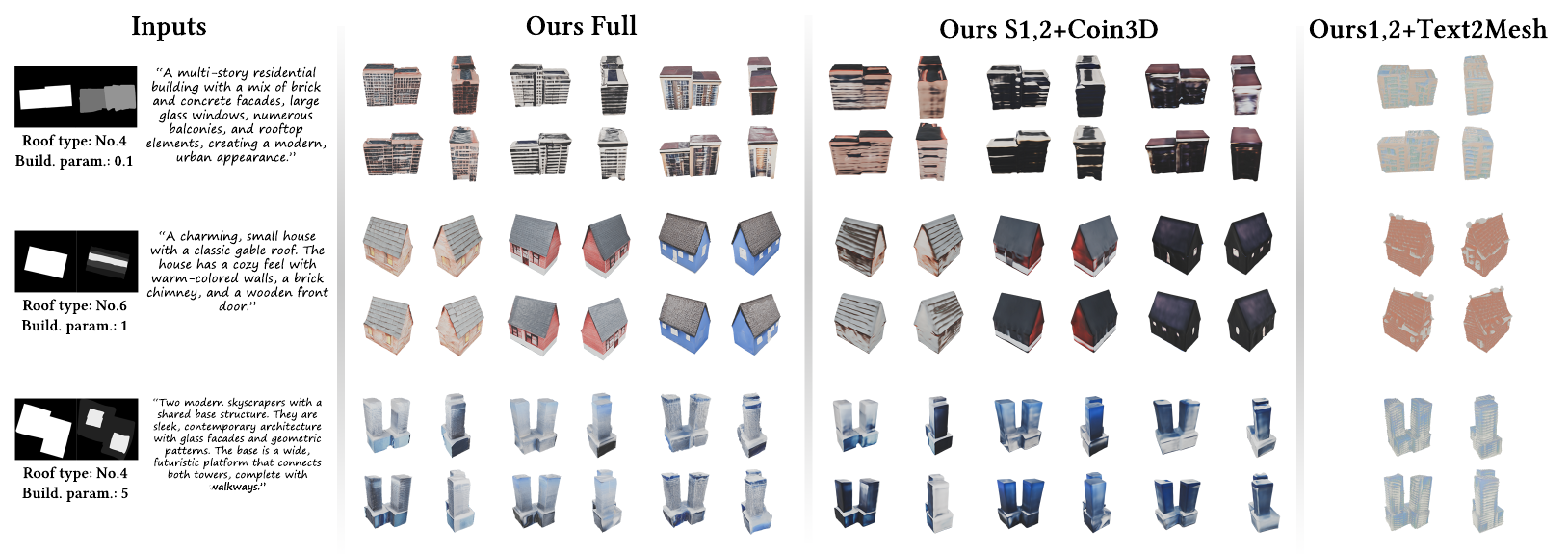}
    \caption{More renderings of the results and corresponding inputs. We continue to provide results with Coin3D and Text2Mesh, as both allow for geometry control. Other text-to-3D and image-to-3D methods are omitted since they lack strict geometric control.}
    \Description{}
    \label{fig:fop1}
\end{figure*}

\begin{figure*}
    \centering
    \includegraphics[width=\linewidth]{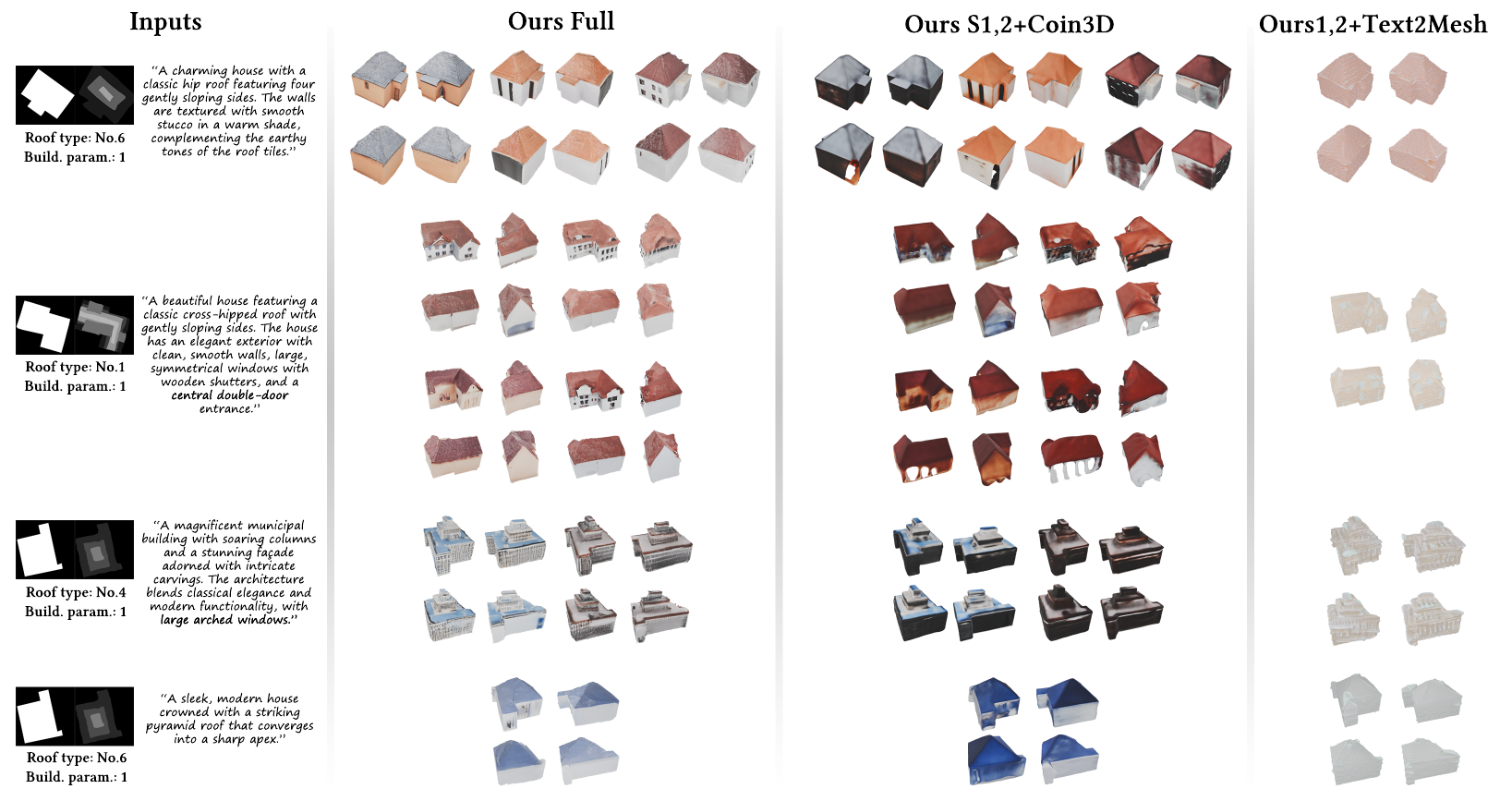}
    \caption{More renderings of the results and corresponding inputs. In these results, stylization begins to suffer from the misalignment discussed in Sec. C.2 of the supplementary material. Despite this, our approach maintains strong structural integrity, whereas Coin3D starts to exhibit holes.}
    \Description{}
    \label{fig:fop2}
\end{figure*}

\begin{figure*}
    \centering
    \includegraphics[width=\linewidth]{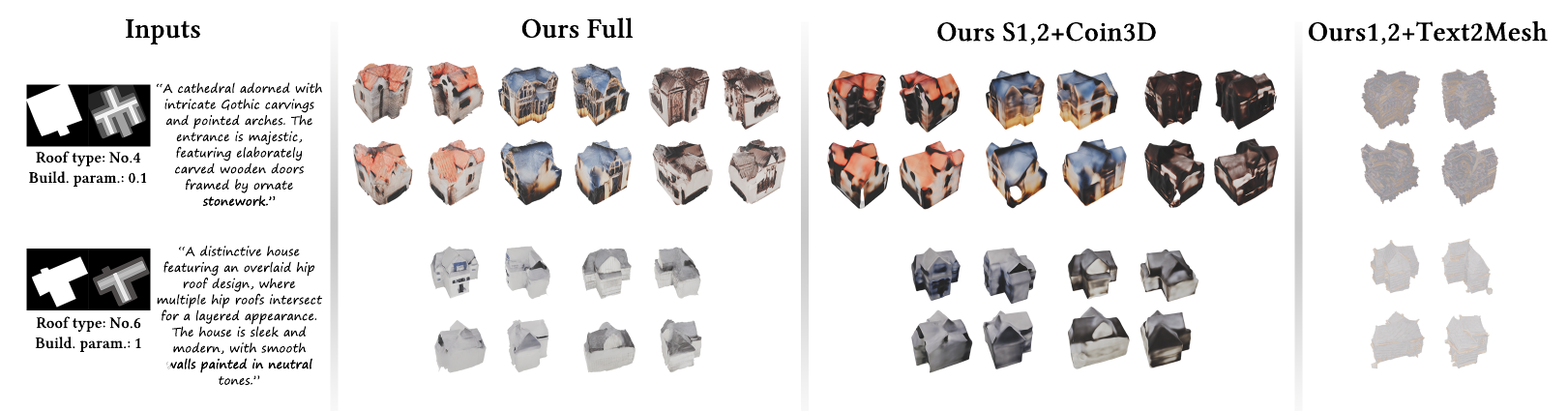}
    \caption{More renderings of the results and corresponding inputs. These results suffer from larger misalignment. Our method still gives more promising results.}
    \Description{}
    \label{fig:fop3}
\end{figure*}

In contrast, the holes observed in the Coins3D outputs (Fig.\ref{fig:fop1} to \ref{fig:fop3}) stem from geometry misalignment. While our results are also affected to some extent, they maintain strong structural integrity and achieve a high IoU with the footprint.

\section{Test Data and More Limitations}
\subsection{Test Data}
For evaluation in Tab.\ref{tab:metrics}, we randomly sampled 20 footprints from the Mesh set of Building3D (excluding the Wireframe set used for S1 training) and added 5 footprints from a city planning platform developed in our previous work. All sketches are drawn manually.

\subsection{Failure Cases}
\paragraph{Failure cases in S1}
When an incorrect or ambiguous type (not the 8 categories in Tab.\ref{tab:roof-prompts}) is provided, the sketch tends to dominate the generation. Since diffusion models learn conditional distributions, this often leads to averaged denoised results of a given sketch.

\paragraph{Failure cases in S2}
As described in Sec.\ref{para:RSR}, We cannot handle non-convex shapes. Another limitation lies in the diversity of modality in the dataset.

\paragraph{Failure cases in S3}
Our appearance generation is based on an MV-diffusion backbone, so any failure in this component directly affects our results. A special case is that the diffusion generates a reasonable appearance, but with poor alignment (middle columns in Fig.\ref{fig:alignment}). To address this, we apply affine transformations and NSF (Sec.\ref{asec:nsf}) to ensure the resulting mesh remains watertight. In contrast, initializing with NeuS does not guarantee watertightness and requires additional training.

% \clearpage

\end{document}